\documentclass[conference]{IEEEtran}
\usepackage{btas}
\usepackage{times}
\usepackage{epsfig}
\usepackage{amsmath}
\usepackage{amssymb}
\usepackage{multirow}
\usepackage[table]{xcolor}
\usepackage{hhline}
\usepackage{graphicx}
\usepackage[table]{xcolor}
\newcommand\ceil[1]{\lceil#1\rceil}

\btasfinalcopy

\ifbtasfinal\fi
\begin{document}

\title{Risk Assessment in the Face-based Watchlist Screening in e-Borders
}

\author{Kenneth Lai, Svetlana N. Yanushkevich, Vlad Shmerko\\
Biometric Technologies Laboratory, Dept. Electrical \& Computer Engineering, University of Calgary\\
{\tt\small \{kelai, syanshk, vshmerko\}@ucalgary.ca}
}

\maketitle

\IEEEoverridecommandlockouts
\IEEEpubid{\begin{minipage}{\textwidth}\ \\[55pt]
		\footnotesize{{\fontfamily{ptm}\selectfont Digital Object Identifier 10.1109/BTAS.2017.8272677 \\978-1-5386-1124-1/17/\$31.00 \copyright 2017 IEEE}}
\end{minipage}}

\begin{abstract}

This paper concerns with facial-based watchlist technology as a component of automated border control machines deployed in e-borders. The key task of the watchlist technology is to mitigate  effects of  mis-identification and impersonation. To address this problem, we developed a novel cost-based model of traveler risk assessment and proved its efficiency via intensive experiments using large-scale facial databases. The results of this study are applicable to any biometric modality to be used in watchlist technology.
 \end{abstract}

\section{Introduction}

The goal of our study is to evaluate the risks of unwanted outcomes of facial-based watchlist screening in Automated Border Control (ABC) machines which are deployed in e-border infrastructure. Varying quality of facial images on the watchlist can result in unintended mis-identification and impersonation, the most dangerous consequence in border crossing. 
  
The two primary functions of the  automated border crossing infrastructure (e-borders) include: 1) authentication of travelers using their appearance and e-passport (most often the e-passport includes a facial template), and 2) search against watchlists \cite{[EC-SmartBorders-2014]}. Both functions must be performed under specific time constraints. Contemporary watchlist screening technology \cite{[US-Department-Justice-Office]} uses only alphanumeric data (such as name, date and place of birth) which are not reliable (can be fabricated). One possible solution for this problem is to integrate biometric traits in watchlists. Biometric technologies have been proven to be effective in Entry-Exit systems for visa visitors \cite{[EC-SmartBorders-2014],[DHS(2010)]}. However, in these applications, only high quality biometric traits are used. In this paper, we study more complicated scenarios of traveler screening in e-borders and risk assessment using watchlist data. 

The main obstacle in achieving a fast and reliable traveler risk assessment is caused by the nature of the watchlist: it usually contains low-quality facial images, including those of non-cooperative individuals, obtained from surveillance \cite{[NIST_Face-In-Video-2017]}. The quality of these images is the main sources of impersonation and mis-identification.

There are various techniques to mitigate this effect; one of them is a mandatory quality management. However, in most cases it is difficult or impossible to control the quality of images on the watchlist.  The use of biometrics may cause additional unwanted effects compared to non-biometric watchlist check, as well as affect the performance of the ABC machines.  The problem is formulated as follows: given a watchlist and a traveler's appearance, assess the risk that an innocent traveler is mis-matched against the watchlist, or a wanted person avoids being matched. These risks are caused by mis-identification and impersonation phenomena.


\section{Related work and contribution}

The central statement of our approach is that the risk assessment in watchlist screening is an inference problem. We utilize Doddington Risk Categorization (DRC) \cite{[kn:Doddington-1998]} in order to create a watchlist risk landscape. Since DRC has an unstable nature, we developed a technique that infers risks using the watchlist landscape and the traveler's information. 

The DRC helps assess the effects of impersonation and mis-identification \cite{[Bustard-Nixon-2013]}; to combat these effects, many studies suggest the use of multiple biometrics \cite{[Poh-Kittler-Bourlai-2010]}. However, in the current ABC machines, the persons of interest are mostly represented by facial traits from both the physical and digital world, and are less likely to be represented by fingerprints or irises \cite{[Best-Rowden-2014]}. 
The recognition process is characterized by False Rejection (FR) (miss-match, or miss-identification) and {False Acceptance (FA)} (false detection, due to impersonation phenomenon) rates, FRR and FAR \cite{[kn:Bolle04]}. 

 An analysis of watchlist operational performance and list size is provided in \cite{[Pato-2010]}; cost-based analysis of the watchlist screening is proposed in \cite{[kn:Bolle04]}. The drawback of this model is that this model is a simplification of the relationship between the cost components, a decision threshold, FRR and FAR.
We consider the use of another metric for risk assessment. We argue that the watchlist risk assessment is a decision inference process, and the cost of mis-identification is only a part of the problem.


In our risk model, a new method of translating recognition scores into risk value is introduced.  This method incorporates the use of four main components: the Doddington Zoo \cite{[kn:Doddington-1998]}, the probability distributions, the similarity/distance concept \cite{[Cha-Measures-2007]}, and the cost as an expected loss \cite{Duda}.  Previous works for each individual component are well-known, however, no existing algorithm combines all four components in the task of modeling risk for watchlist screening.



\section{Watchlist risk assessment model}\label{sec:Extending}


{Doddington Risk Categorization (DRC)} is defined as a classification of individuals, using their biometric traits, into four types \cite{[kn:Doddington-1998]}: 
I (`sheep'), who  are recognized normally and have high genuine scores and low impostor scores;
II (`goats'), who  are hard to recognize and have low genuine scores;
III (`wolves'), who are good at impersonating and have high impostor scores; and
IV (`lambs'), who are easy to impersonate and have high genuine and impostor scores.

DRC utilizes the two types of match scores: \emph{genuine} and \emph{impostor}. They are the results of comparing two biometric samples that belong to the same and different individual, respectively. A technique for analysis of scores is explained, in particular, in \cite{[Sgroi-2015]}.

In order to translate both the genuine and imposter scores into DRC, a ranking of the scores is required. `Goats' are selected by finding a certain number (2.5\% of population) of individuals that contain low genuine scores, as they do not match well against themselves. 
`Wolves' and `lambs' are an impostor/victim pair that are found by identifying a certain number (2.5\% of population) of individuals with high impostor scores.`Sheep' are individuals that neither belong to the 'goat' nor the 'wolf' categories \cite{[kn:RossRattani-2009]}.
The reason for choosing the statistically significant 2.5\% of the population is that it represents two standard deviations from the averaged recognition scores. This is normally accepted in the matching techniques \cite{[kn:Doddington-1998]}.
{Watchlist landscape} is defined as the biometric watchlist content over the DRC.

Two biometric samples of the same identity are rarely quite similar due to a number of factors. To tolerate such variations, a distance or similarity measure between these samples is needed. The choice of such measure depends on the representation of samples; in this paper, we use probability distributions.

In our model, risk $R$ is defined as an expected loss \cite{Duda}. Based on this concept, we propose the following formula to estimate the risk. The total risk of traveler mis-identification using watchlist screening is defined as:
\vspace{-3mm}
\begin{equation}\label{eq:risk}
	R=1-\sum^{t}_{j=1}\lambda(j)D(j)
\end{equation}
where $t$ is the number of DRCs, $\lambda(j)$ is the loss/cost associated with the risk category $j$ and $D(j)$ is the dissimilarity to category $j$. 


We assume that
1) 'Sheep' have low risk because they do not cause many mis-identifications or rejections, $\lambda(\text{`sheep'})=0.1$;
2) 'Goats' have medium risk because they cause false rejects which impacts the throughput of the system, $\lambda(`goat')=0.3$; and
3) 'Wolves'/`Lambs' (`W'/`L') have high risk because they impact false accepts which jeopardizes the security of the system, $\lambda(`wolf'/`lamb')=0.6$. Therefore, the risk level is assigned as follows:
\textbf{Low risk} is assigned to travelers who do not impact the system or their impact is negligible. 
\textbf{Medium risk} is assigned to travelers who are rejected by the system, therefore costing additional resources to redress (attempts to access multiple times or technician assistance). 
\textbf{High risk} is assigned to travelers who have gained access to system without having the required credentials.

At each phase of the traveler risk assessment, uncertainty content should be evaluated. In our study, uncertainty is represented by \emph{probability distributions}. For measurement and comparisons over this data structure landscape, we used different metrics \cite{[Cha-Measures-2007],[Jousselme-2012]}.


Since these similarity measurements only reports the degree of similarity, a classifier is required to interpret the degree of likeliness to a specific DRC. It utilizes the similarity measurements computed from the selected Doddington categories as the features for finding the corresponding DRC for each person.  In order to find the best performing metric for our risk assessment model, we compared the results of two different classifiers: 
 $(a)$ Support Vector Machines (SVM) \cite{Duda} and
 $(b)$ minimum similarity, or distance, function:
\begin{equation}\label{eq:min-sim/dist-function}
	\min (M_{i,\text{`goat'}}, M_{i,\text{`wolf'/`lamb'}}, M_{i,\text{`sheep'}})
\end{equation}
where $M_{i,\text{`X'}}$ denotes the similarity between probability distributions for a Doddington risk category $X$ in metric $i$. 
	
The minimum similarity function is chosen as a classifier because it does not require any training. The other type of classifier, SVM, is  a supervised learning model which uses the training data to find the boundaries separating the training data into classes.
Consider, for example, a case for Subject A when the Euclidean distance of the subject to the `goat', `wolf'/`lamb' and `sheep' categories is measured as 0.1, 0.2 and 0.7. When using $\min()$ function as a classifier, 0.1 values means that a `goat' category is chosen as the predicted class for Subject A. If Subject A's real category is also a `goat', then the sensitivity is 100\%; however, if Subject A's real category is not a `goat', then the sensitivity is 0\%.

\section{Extension of the watchlist technology}\label{sec:Extension}


For the watchlist technology, there are two kinds of risks: the watchlist itself (as a data structure that stores data on multiple persons of interest) and the watchlist screening process. These different kinds of risks address two approaches, or levels of their control. {Risk level control} is defined by the following scenarios of the watchlist application.

 {Level-I} (host system): Given the watchlist, its potential risk for an arbitrary traveler is defined by the distribution of data structure in an appropriate metric. 
 
 {Level-II} (authentication and risk assessment station): Given a traveler, the risk of his/her watchlist screening is defined with respect to the potential risks of this watchlist.
Hence,  at least the following phases should be added to the watchlist technology:

\textbf{Phase I: DRC detection.} Nature of the DRC is unstable and depends on many uncontrollable factors. There are several kinds of DRC detectors, as well as various metrics for measurement of similarities of DRC \cite{[Lai-Yanushkevich-2017]}.

\textbf{Phase II: Watchlist landscape monitoring.} Watchlist is a dynamical structure characterized by the number and type of identities, costs of various unwanted effects and the risks that  vary over the time. These characteristics should be periodically updated, and the control parameters must be adjusted. 

\textbf{Phase III: Cost specification.} Generic specification of the cost of mis-identifying a person of interest and the cost of mis-matching a regular traveler. These costs will depend on the national security policy and the current security indicators. 

\textbf{Phase IV: Automated quality control monitoring} such as \cite{[Abaza-2014]}, should be used after each watchlist update.  


\section{Experimental results}\label{sec:Experiment}
In order to confirm the efficiency and rationality of the proposed risk model based on the multi-metric assessment and cost control screening, an intensive experimental study was conducted.  The experimental study involved the four main phases described  in Section 4, namely,  a) DRC detection, 
 b) Watchlist landscape monitoring, 
 c) Cost specification, and 
 d) Automated quality control monitoring.

The goal of this experimental study is to prove the utility of the proposed watchlist technology extension which includes the four phases (Section \ref{sec:Extension}).    

\subsection{Databases and tools}

The FRGC \cite{FRGC} and  LFW \cite{LFWTech} databases were used for experiments. Out of the FRGC V2.0 database's 50 000 recordings with 4007 subject sessions, we used 30 000 images of 487 subjects that represents both the controlled and uncontrolled illumination settings. Also, we selected 13 000 images from the LFW database consisting of unconstrained web photos of 5749 different subjects. We used a commercial face recognition tool Verilook 
which is based on neural network approach and is one of the most accurate tools as per Facial Recognition Vendor Test 2013 NIST competition  \cite{frvt2013}. It is well known that the DRC varies when different recognition algorithm is applied  \cite{[kn:Bolle04],[Bustard-Nixon-2013],[Poh-Kittler-Bourlai-2010],[Sgroi-2015]}. However, the DRC  phenomenon  is independent of the size of database: given a traveler and a watchlist, a DRC can result in a decision such as false reject or false accept. The proposed approach can be used for any biometric-enabled watchlist, regardless of size and modality.

\subsection{Watchlist landscape monitoring}

The watchlist is a dynamic data structure because data can be added (\eg, new facial images), replaced (\eg, by better quality facial images), or deleted (in a redress mechanism). These changes can impact the Doddington landscape.
 {Watchlist landscape monitoring} is defined as a re-calculation of the DRC after any biometric trait data changes.

In the scenario of DRC, the probability of occurrence can be predicted based on the scores resulted from facial recognition. Our method of prediction uses an approach of comparing probability distributions as a tool to measure similarity. 
In this experiment, the database is separated into disjointed sets based on the quality (high, low, various) and categories (`goats', `wolves'/`lambs', and `sheep') to a total of 9 subsets.  In addition, when analyzing specific subjects, their contributions to the DRC distributions are discarded. Therefore, there is no prior assumption or relationship between the subjects and the DRC.

\begin{figure*}[!hbt]
	\begin{tabular}{cc}
		\centering
		\hspace{-2mm}
		\begin{parbox}[h]{0.5\linewidth}{\centering 
		\includegraphics[width=0.44\textwidth]{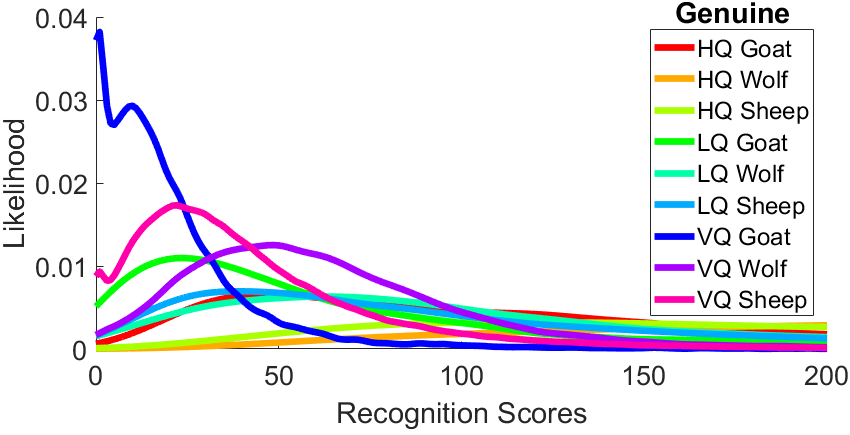}
		}\end{parbox}
		&\hspace{-5mm}
		\begin{parbox}[h]{0.5\linewidth}{\centering 
		\includegraphics[width=0.44\textwidth]{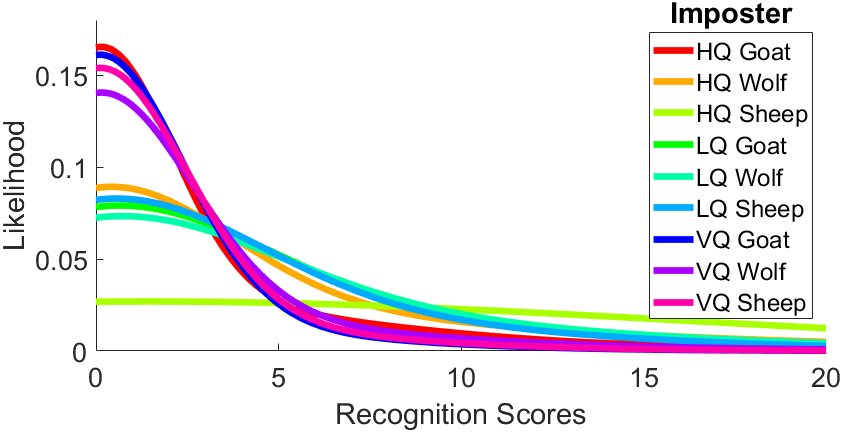}
		}\end{parbox}
	\end{tabular}
	\caption{Watchlist landscape monitoring: The probability distribution of each DRC with different types of image quality comparisons (HQ, LQ, VQ). Left plane: Genuine score distribution; Right plane: Imposter score distribution.}
	\label{fig:q-doddington-curve}
\end{figure*}

Phase II ``Watchlist landscape monitoring'' is introduced in Fig.~\ref{fig:q-doddington-curve} as the score distribution for each Doddington risk category (`goats', `sheep', `wolves'/`lambs') at different image qualities (high, low, various). In the FRGC database, there are two main groups of images: the controlled and the uncontrolled ones. Both groups contain full frontal faces; the controlled images were taken in a fixed and well-lit environment, whereas uncontrolled ones were taken in complex environments (such as hallways). 
In our experiments, these uncontrolled images  caused low recognition scores and, thus, were treated as “low quality” ones. It should be noted that more details on image quality control can be found in \cite{alonso2012quality,grother2007performance}. We performed three types of image comparison: high vs high (high or HQ), low vs low (low or LQ), high vs low and low vs high combined (various or VQ), respectively.
In our experiments, we followed the generally accepted DRC technique \cite{[kn:Doddington-1998],frvt2013,[Poh-Kittler-Bourlai-2010],[Sgroi-2015]} to represent the likelihood of the face matching score derived for each DRC with respect to the score bin (Fig. 2). The distributions of the genuine and imposter scores are calculated by creating a histogram where a Gaussian function is used to describe the probability distribution. To create such distribution, all pair of faces are considered within the specified DRCs. For example, the genuine HQ Goat includes all image comparisons between faces that are genuine, HQ, and `goats'. Fig. 2 illustrates the genuine and imposter score distribution for each DRC (‘goats', ‘sheep', ‘wolves'/’lambs') using all possible comparisons of images of different quality (HQ, LQ, VQ) for the FRGC database.''


The population percentage for each DRC is shown in Table \ref{tab:entireZoo}. The distribution of DRC is derived by counting the subjects with low genuine or high imposter scores. This approach is similar to the one defined in \cite{[kn:RossRattani-2009]} which locates the DRC by ranking the order of genuine and imposter scores. 

\begin{table}[!htb]
\centering
\caption{Probability distributions of DRC for FRGC and LFW databases }\label{tab:entireZoo}\vspace{3mm}
\begin{small}
\begin{tabular}{|l||r|r|r|r|}
		\hline
		 Database' & `Goat' & `Lamb' & `Wolf' & `Sheep' \\
		\hline
		\hline
		FRGC & 0.0264& \multicolumn{2}{c|}{0.0264} & 0.9472\\
		\hline
		LFW & 0.0253& \multicolumn{2}{c|}{0.0253} & 0.9506\\
		\hline
\end{tabular}
\end{small}
\end{table}

In this paper, 2.5\% of the population with the lowest genuine scores are chosen as `goat', another 2.5\% of the population with the highest imposter scores are selected as `wolf'/`lamb', and the remaining population are placed into `sheep' category. Since the population is 568 for the selected subset of the FRGC database, the percentile population can be calculated as $\ceil{\text{Percentile} \times \text{Population}}$, 2.5\% represents 15 ($\ceil{0.025 \times 568}$) subjects.
	
We observe that the quality of the images is the main factor affecting the risks. In Fig.~\ref{fig:q-doddington-curve}, the genuine HQ curves demonstrate a more even distribution, whereas genuine VQ curves represent a much more concentrated distribution.
 

\subsection{Performance evaluation}

Performance of the watchlist screening includes many parameters, such as the depth of social embedding, types of identity traits, FAR, FRR, infrastructure features and redress metric \cite{[Lai-Yanushkevich-2017]}. In this paper, the watchlist performance addresses only the risks assessment of traveler screening.
{Watchlist performance} is defined as the efficiency of risk screening over the watchlist landscape given the traveler's DRC.
	
In this Section, we introduce the following mechanism for the efficiency evaluation:
1)	Graphical representation of risk scenarios,
2)	Comparison of risk scenarios in appropriate metric(s) and
3)	Risk assessment and decision-making.	
	

Given the DRC for each person, we performed a leave-one-out cross validation to estimate the performance of each classifier at different levels of image quality as reported in Table \ref{tab:svm}. 
The data used to generate Table \ref{tab:svm} is derived from the results of all image comparisons, therefore, Table \ref{tab:svm} reports the overall summarized sensitivity of using the selected metrics. Since there are genuine and imposter distributions for each DRC, the results are divided into genuine and imposter columns.

Each row in Table \ref{tab:svm} signifies the metric used for comparing the different probability distributions; the columns of the table indicate the quality of images used for comparison (high, low and various). The numbers in the table denotes the sensitivity given the row (metric) and column (quality). The highlighted cells indicate the highest sensitivity for each column. 

\begin{table*}[!t]
	\centering
		\caption{DRC classification in various metrics}\vspace{3mm}
				\begin{footnotesize}
		\begin{tabular}{|l||r|r|r|r|r|r||r|r|r|r|r|r|}
		\hline
			&	\multicolumn{6}{c||}{\small Support Vector Machine}&	\multicolumn{6}{c|}{\small Minimum Similarity/Distance Function}	\\
			\hline
	&	\multicolumn{3}{c|}{\small Genuine}&	\multicolumn{3}{c||}{\small Imposter}&	\multicolumn{3}{c|}{\small Genuine}&	\multicolumn{3}{c|}{\small Imposter}	\\
	\hline
	&	High	&	Low	&	Various	&	High	&	Low	&	Various	&	High	&	Low	&	Various	&	High	&	Low	&	Various\\
	\hline
	\hline
Euclidean	&	0.69	&	0.60	&	\textbf{0.80}	&	0.67	&	0.33	&	0.33	&	0.82	&	0.64	&	\textbf{0.80}	&	0.44	&	0.40	&	0.49	\\
City Block	&	0.69	&	0.62	&	\textbf{0.80}	&	\textbf{0.69}	&	0.44	&	0.51	&	\textbf{0.84}	&	0.64	&	0.78	&	0.51	&	0.40	&	0.49	\\
Chebyshev	&	0.73	&	0.58	&	0.76	&	0.64	&	0.51	&	0.33	&	\textbf{0.84}	&	0.64	&	0.69	&	0.29	&	0.38	&	0.49	\\
\hline																									
Sorensen	&	0.69	&	0.60	&	\textbf{0.80}	&	\textbf{0.69}	&	0.44	&	0.51	&	\textbf{0.84}	&	0.69	&	0.78	&	0.51	&	0.40	&	0.49	\\
Canberra	&	0.60	&	0.53	&	0.64	&	0.67	&	0.47	&	0.60	&	0.58	&	0.33	&	0.53	&	0.51	&	0.49	&	\textbf{0.64}	\\
Lorentzian	&	0.69	&	0.62	&	\textbf{0.80}	&	\textbf{0.69}	&	0.44	&	0.51	&	\textbf{0.84}	&	0.64	&	0.78	&	0.53	&	0.40	&	0.49	\\
\hline																									
Wave Hedges	&	0.62	&	0.58	&	0.64	&	0.67	&	0.49	&	\textbf{0.62}	&	0.29	&	0.33	&	0.33	&	0.33	&	0.33	&	0.33	\\
Czekanowski	&	0.69	&	0.60	&	\textbf{0.80}	&	\textbf{0.69}	&	0.44	&	0.51	&	\textbf{0.84}	&	0.69	&	0.78	&	0.51	&	0.40	&	0.49	\\
Kulczynski s	&	\textbf{0.87}	&	0.64	&	0.78	&	0.58	&	0.49	&	0.51	&	0.02	&	0.02	&	0.00	&	0.31	&	0.24	&	0.16	\\
\hline																									
Harmonic means	&	0.80	&	0.49	&	0.69	&	\textbf{0.69}	&	0.51	&	\textbf{0.62}	&	0.02	&	0.02	&	0.00	&	0.31	&	0.29	&	0.20	\\
Kumar-Hassebrook	&	0.84	&	\textbf{0.67}	&	0.78	&	0.67	&	0.53	&	0.60	&	0.02	&	0.02	&	0.00	&	0.31	&	0.29	&	0.18	\\
Jaccard	&	0.69	&	0.64	&	0.76	&	0.67	&	0.36	&	0.51	&	0.80	&	0.64	&	\textbf{0.80}	&	0.47	&	0.40	&	0.49	\\
\hline																									
Hellinger	&	0.71	&	0.58	&	0.73	&	\textbf{0.69}	&	0.56	&	0.58	&	0.82	&	\textbf{0.73}	&	0.76	&	\textbf{0.60}	&	0.44	&	0.58	\\
Matusita	&	0.71	&	0.58	&	0.73	&	\textbf{0.69}	&	0.56	&	0.58	&	0.82	&	\textbf{0.73}	&	0.76	&	\textbf{0.60}	&	0.44	&	0.58	\\
Squared-Chord	&	0.67	&	0.64	&	0.71	&	\textbf{0.69}	&	0.53	&	0.58	&	0.82	&	\textbf{0.73}	&	0.76	&	\textbf{0.60}	&	0.44	&	0.58	\\
\hline																									
Squared Euclidean	&	0.67	&	0.62	&	0.76	&	0.67	&	0.36	&	0.51	&	0.82	&	0.64	&	\textbf{0.80}	&	0.44	&	0.40	&	0.49	\\
Squared	&	0.67	&	0.62	&	0.73	&	0.67	&	0.56	&	0.58	&	0.82	&	0.67	&	0.76	&	\textbf{0.60}	&	0.47	&	0.58	\\
Clark	&	0.62	&	0.53	&	0.62	&	\textbf{0.69}	&	0.47	&	0.58	&	0.53	&	0.33	&	0.53	&	0.58	&	0.47	&	0.62	\\
\hline																									
Kullback-Leibler	&	0.56	&	0.33	&	0.51	&	0.67	&	0.51	&	0.60	&	0.38	&	0.38	&	0.38	&	0.33	&	\textbf{0.62}	&	0.58	\\
K Divergence	&	0.67	&	0.58	&	0.69	&	0.67	&	\textbf{0.58}	&	0.53	&	0.80	&	0.69	&	0.76	&	0.40	&	0.53	&	0.47	\\
Jensen-Shannon	&	0.67	&	0.62	&	0.71	&	0.67	&	0.56	&	0.58	&	0.82	&	\textbf{0.73}	&	0.76	&	\textbf{0.60}	&	0.44	&	0.58	\\
		\hline
		\end{tabular}
			\end{footnotesize}
		\label{tab:svm}
\end{table*}

Table \ref{tab:svm} confirms that each metric generally performs very well with genuine comparisons regardless of the classifier. 
The results of the genuine portion show  higher sensitivity when compared to the imposter. This indicates that the genuine comparisons can be used as an indicator of the ‘wolf'/’lamb' category.  For example, using Euclidean distance and SVM, the genuine VQ results in accuracy of 0.8 whereas for imposter VQ generates accuracy of 0.33.

The best performing metric for the high quality comparison and the overall highest sensitivity is the Kulczysnki $s$ metric. Other metrics causes the performance ranges from 0.60 to 0.84 for various image quality comparisons.

Using the risk estimation (Equation \ref{eq:risk}) with the results from a selected metric and the pre-determined loss values, we are able to determine the risk associated with each individual.
Table \ref{tab:metricex} reports the results of risk assessment for three different subjects, each from a different DRC, given no classifier (R1), the minimum function (R2), and the SVM classifier (R3) for the Euclidean metric. Evaluation of each of the three subjects and their corresponding DRC are performed by comparing each image's score distribution and the DRC distribution.

In Table \ref{tab:metricex}, each row represents the quality of compared images (H for high, L for low and V for various), each column indicates the DRC and each cell reports the distance between  the probability distributions. For each cell, the value is normalized to be between 0 and 1, where 0 indicates identical images and 1 implies  absolutely different ones. The \textbf{bold} (red), \underline{underlined} (blue) values indicates the output of the min function and SVM classifiers, respectively. The \textbf{\underline {bold and underlined}} (green) indicates the value which is identical for both   classifiers.

Consider the scenario of the watchlist screening in which the traveler is pre-determined to be in the category `goat' (subject 4315 in the database) and only the high quality image comparisons are used. According to Equation \ref{eq:risk}, the risk values of watchlist screening of genuine and imposter score when no classifiers are used are:
$R1_{G}=1-(0.3 \times 0.18+0.6 \times 0.78+0.1 \times 0.61)={0.42}$ and  
$R1_{I}=1-(0.3 \times 0.48+0.6 \times 0.16+0.1 \times 0.86)={0.67}$, respectively. 
This traveler yields risk values 0.36 to 0.42 and 0.48 to 0.67 for genuine and imposter comparisons, respectively and the average is 0.485. Repeating the same procedure using the classifiers produces the following results: $R2_{G}=0.30$, $R2_{I}=0.60$, $R3_{G}=0.30$, and $R3_{I}=0.30$. 
In security terms, these risk means that the traveler has a risk level between medium and high for no classifier. For the minimum function classifier, it indicates minimal risk, and for the SVM classifier, the risk level is determined to be medium.
 
\begin{table}[!t]
\centering
\caption{Similarity Measurement using the Euclidean metric} \label{tab:metricex}
	\begin{footnotesize}
		\begin{tabular}{|l|l|l||r|r|r||r|r|r|}
			\hline
			& & Q. & Goat & W/L & Sheep & R1 & R2 & R3\\
			\hhline{=========}
			\parbox[t]{2mm}{\multirow{6}{*}{\rotatebox[origin=c]{90}{4315	(`Goat')}}}	&	\parbox[t]{2mm}{\multirow{3}{*}{\rotatebox[origin=c]{90}{Gen.}}}	&H	&	\cellcolor{green!25}\textbf{\underline{0.18}}	&	0.78	&	0.61	&	0.42	&	0.30	&	0.30	\\ 
			&		&	L	&	0.65	&	0.64	&	\cellcolor{green!25}\textbf{\underline{0.41}}	&	0.38	&	0.10	&	0.10	\\
			&		&	F	&	\cellcolor{green!25}\textbf{\underline{0.32}}	&	0.83	&	0.46	&	0.36	&	0.30	&	0.30	\\
			\hhline{~--------}				
			&	\parbox[t]{2mm}{\multirow{3}{*}{\rotatebox[origin=c]{90}{Imp.}}}	&	H	&	\cellcolor{blue!25}\underline{0.48}	&	\cellcolor{red!25}\textbf{0.16}	&	0.86	&	0.67	&	0.60	&	0.30	\\
			&		&	L	&	0.59	&	\cellcolor{green!25}\textbf{\underline{0.47}}	&	0.66	&	0.48	&	0.60	&	0.60	\\
			&		&	F	&	0.82	&	\cellcolor{green!25}\textbf{\underline{0.16}}	&	0.55	&	0.60	&	0.60	&	0.60	\\
			\hhline{---------}
			\parbox[t]{2mm}{\multirow{6}{*}{\rotatebox[origin=c]{90}{4202	(`W'/`L')}}}	&	\parbox[t]{2mm}{\multirow{3}{*}{\rotatebox[origin=c]{90}{Gen.}}}	&H	&	0.92	&	\cellcolor{green!25}\textbf{\underline{0.14}}	&	0.37	&	0.60	&	0.60	&	0.60	\\
			&		&	L	&	\cellcolor{blue!25}\underline{0.55}	&	0.69	&	\cellcolor{red!25}\textbf{0.47}	&	0.37	&	0.10	&	0.30	\\
			&		&	F	&	0.93	&	\cellcolor{blue!25}\underline{0.30}	&	\cellcolor{red!25}\textbf{0.22}	&	0.52	&	0.10	&	0.60	\\
			\hhline{~--------}													
			&	\parbox[t]{2mm}{\multirow{3}{*}{\rotatebox[origin=c]{90}{Imp.}}}	&	H	&	0.80	&	\cellcolor{green!25}\textbf{\underline{0.19}}	&	0.57	&	0.59	&	0.60	&	0.60	\\
			&		&	L	&	0.50	&	\cellcolor{green!25}\textbf{\underline{0.22}}	&	0.84	&	0.63	&	0.60	&	0.60	\\
			&		&	F	&	0.74	&	\cellcolor{green!25}\textbf{\underline{0.31}}	&	0.60	&	0.53	&	0.60	&	0.60	\\
			\hhline{---------}
			\parbox[t]{2mm}{\multirow{6}{*}{\rotatebox[origin=c]{90}{2463	(`Sheep')}}}	&	\parbox[t]{2mm}{\multirow{3}{*}{\rotatebox[origin=c]{90}{Gen.}}}	&H	&	0.77	&	0.57	&	\cellcolor{green!25}\textbf{\underline{0.27}}	&	0.40	&	0.10	&	0.10	\\
			&		&	L	&	0.86	&	\cellcolor{blue!25}\underline{0.47}	&	\cellcolor{red!25}\textbf{0.20}	&	0.44	&	0.10	&	0.60	\\
			&		&	F	&	0.59	&	0.75	&	\cellcolor{green!25}\textbf{\underline{0.30}}	&	0.34	&	0.10	&	0.10	\\
			\hhline{~--------}
			&	\parbox[t]{2mm}{\multirow{3}{*}{\rotatebox[origin=c]{90}{Imp.}}}	&	H	&	\cellcolor{green!25}\textbf{\underline{0.05}}	&	0.43	&	0.90	&	0.64	&	0.30	&	0.30	\\
			&		&	L	&	\cellcolor{blue!25}\underline{0.52}	&	0.76	&	\cellcolor{red!25}\textbf{0.38}	&	0.35	&	0.10	&	0.30	\\
			&		&	F	&	\cellcolor{green!25}\textbf{\underline{0.47}}	&	0.69	&	0.55	&	0.39	&	0.30	&	0.30	\\
			\hline
		\end{tabular}
	\end{footnotesize}
\end{table}

When no classifiers (R1) are used, the genuine comparison (Table \ref{tab:metricex}) indicates that the `wolf'/`lamb' travelers  yield on average the highest value (0.50) of risk, while the `sheep' (0.39) and 'goat' (0.39)  provide a much lower risk value. When incorporating the use of both genuine and imposter comparisons, we can see the level of risk increasing in the following order: `wolf'/`lamb', `goat' and `sheep' (0.54, 0.49, 0.43). 

Since both classifiers select only one DRC as the final choice, the resulting risk calculations (R2 and R3) reports the loss value corresponding to the selected category. When averaging the results of the minimum function classifier, `sheep'  are easily identified by their low R2 values (0.17) compared to greater values (0.42 and 0.43) for the `goat' and `wolf'/`lamb'. However, since the `goat' and `wolf'/`lamb' have similar risk values, it is difficult to distinguish the two categories. In case of SVM, the averaged results show three distinct separations, 0.37, 0.55 and 0.28 to their respective `goat', `wolf'/`lamb' and `sheep' categories; it allows for better classification.

The reported risk values provide better identification of `goat' and `wolf'/`lamb', due the larger difference in the categories' risk values. Specifically, SVM indicates a better separation between each individual DRC as opposed to the minimum function which can only separate `sheep' from either `goat' or `wolf'/`lamb' categories.

\section{Conclusions}\label{sec:Conclusion}


The reported results lead to the following conclusions.
1) The proposed watchlist risk model  is well suited for analysis of unwanted effects, and the selected classifiers  performed well for the DRC task.
2) The suggested experimental protocol  is well suited for benchmarking and  comparison of various risk inference techniques. 
3) Risk can be predicted using the watchlist landscape monitoring accordingly to the additional design/control phases proposed for the watchlist screening.
Finally, while our study only highlights the existing problems of biometric-enabled watchlist technology, we state that the future lies in developing a powerful inference engine for dealing with high-conflicting information and uncertainty. We are currently working on improving the results using the predictors proposed in \cite{scheirer2012learning}.

\section*{Acknowledgments}
\vspace{-2mm}
\begin{small}
This project was partially supported by  Natural Sciences and Engineering Research Council of Canada (NSERC), grant ``Biometric intelligent interfaces''; the Government of the Province of Alberta, ASRIF grant and Queen Elizabeth II Scholarship; and Defense Research and Development Canada.
\end{small}

{\small
\bibliographystyle{ieee}
\bibliography{egbib}

\begin{thebibliography}{10}\itemsep=-1pt

\bibitem{[Abaza-2014]}
A.~Abaza, M.~A. Harrison, T.~Boulai, and A.~Ross.
\newblock Design and evaluation of photometric image quality measures for
  effective face recognition.
\newblock {\em IET Biometrics}, 3(4):314--324, 2014.

\bibitem{alonso2012quality}
F.~Alonso-Fernandez, J.~Fierrez, and J.~Ortega-Garcia.
\newblock Quality measures in biometric systems.
\newblock {\em IEEE Security \& Privacy}, 10(6):52--62, 2012.

\bibitem{[Best-Rowden-2014]}
L.~Best-Rowden, H.~Han, and C.~Otto~et al.
\newblock Unconstrained face recognition: Identifying a person of interest from
  a media collection.
\newblock {\em IEEE Trans. Inf. Forensics and Security}, 9(12):2144--2157,
  2014.

\bibitem{[kn:Bolle04]}
R.~M. Bolle, J.~Connell, S.~Pankanti, N.~K. Ratha, and A.~W. Senior.
\newblock {\em Guide to biometrics}.
\newblock Springer, 2013.

\bibitem{[Bustard-Nixon-2013]}
J.~Bustard, J.~Carter, and M.~Nixon.
\newblock Targeted impersonation as a tool for the detection of biometric
  system vulnerabilities.
\newblock In {\em Proc. IEEE 6th Int. Conf. on Biometrics: Theory, Applications
  and Systems}, pages 1--6, 2013.

\bibitem{[Cha-Measures-2007]}
S.-H. Cha.
\newblock Comprehensive survey on distance/similarity measures between
  probability density functions.
\newblock {\em Int. J. Math. Models and Methods in Applied Sci.}, 1(2):1, 2007.

\bibitem{[kn:Doddington-1998]}
G.~Doddington, W.~Liggett, and A.~Martin, et~al.
\newblock Sheep, goats, lambs and wolves: A statistical analysis of speaker
  performance in the {NIST} 1998 speaker recognition evaluation.
\newblock Technical report, 1998.

\bibitem{Duda}
R.~O. Duda, P.~E. Hart, and D.~G. Stork.
\newblock {\em Pattern classification}.
\newblock John Wiley \& Sons, 2012.

\bibitem{[EC-SmartBorders-2014]}
{EU European Commission B-1049}.
\newblock {Technical Study on Smart Borders}, 2014.

\bibitem{frvt2013}
P.~Grother and M.~Ngan.
\newblock Face recognition vendor test ({FRVT}) performance of face
  identification algorithms.
\newblock Technical report, NIST, 2013.

\bibitem{[NIST_Face-In-Video-2017]}
P.~Grother, G.~Quinn, and M.~Ngan.
\newblock {Face In Video Evaluation (FIVE) Face Recognition of Non-Cooperative
  Subjects}.
\newblock Technical Report 8173, NIST, 2017.

\bibitem{grother2007performance}
P.~Grother and E.~Tabassi.
\newblock Performance of biometric quality measures.
\newblock {\em IEEE Trans. on pattern analysis and machine intelligence},
  29(4):531--543, 2007.

\bibitem{LFWTech}
G.~B. Huang, M.~Ramesh, T.~Berg, and E.~Learned-Miller.
\newblock Labeled faces in the {W}ild: A database for studying face recognition
  in unconstrained environments.
\newblock Technical Report 07-49, University of Massachusetts, Amherst, 2007.

\bibitem{[Jousselme-2012]}
A.-L. Jousselme and P.~Maupin.
\newblock Distances in evidence theory: Comprehensive survey and
  generalizations.
\newblock {\em Int. J. Approximate Reasoning}, 53(2):118--145, 2012.

\bibitem{[Lai-Yanushkevich-2017]}
K.~Lai, S.~N. Yanushkevich, V.~P. Shmerko, and S.~C. Eastwood.
\newblock Bridging the gap between forensics and biometric-enabled watchlists
  for e-borders.
\newblock {\em IEEE Comput. Intell. Magazine}, 12(1):16--28, 2017.

\bibitem{[Pato-2010]}
J.~N. Pato and L.~I. Millett, editors.
\newblock {\em Biometric Recognition: Challenges and Opportunities}.
\newblock National Academies Press, 2010.

\bibitem{FRGC}
P.~J. Phillips, P.~J. Flynn, and T.~Scruggs, et~al.
\newblock Overview of the face recognition grand challenge.
\newblock In {\em Proc. IEEE Computer Society Conf. on Computer Vision and
  Pattern Recognition}, pages 947--954, 2005.

\bibitem{[Poh-Kittler-Bourlai-2010]}
N.~Poh, J.~Kittler, and T.~Bourlai.
\newblock Quality-based score normalization with device qualitative information
  for multimodal biometric fusion.
\newblock {\em IEEE Trans. Syst., Man, and Cybern. -- Part A: Syst. and
  Humans}, 40(3):539--554, 2010.

\bibitem{[kn:RossRattani-2009]}
A.~Ross, A.~Rattani, and M.~Tistarelli.
\newblock Exploiting the {Do}ddington zoo effect in biometric fusion.
\newblock In {\em IEEE 3rd Int. Conf. on Biometrics: Theory, Applications, and
  Syst.}, 2009.

\bibitem{scheirer2012learning}
W.~J. Scheirer, A.~de~Rezende~Rocha, J.~Parris, and T.~E. Boult.
\newblock Learning for meta-recognition.
\newblock {\em IEEE Transactions on Information Forensics and Security},
  7(4):1214--1224, 2012.

\bibitem{[Sgroi-2015]}
A.~Sgroi, P.~J. Flynn, K.~Bowyer, and P.~J. Phillips.
\newblock Strong, neutral, or weak: Exploring the impostor score distribution.
\newblock {\em IEEE Inf. Forensic and Security}, 10(6):1207--1220, 2015.

\bibitem{[DHS(2010)]}
{U.S. Department of Homeland Security}.
\newblock {Biometric Standards Requirements for US-VISIT, Version 1.0}, 2010.

\bibitem{[US-Department-Justice-Office]}
{U.S. Department of Justice Office of the Inspector General}.
\newblock {Follow-Up Audit of the Terrorist Screening Center}, 2007.

\end{thebibliography}
}

\end{document}